\PassOptionsToPackage{hyphens}{url}
\documentclass{fairmeta}

\definecolor{rageR}{HTML}{3478B8} 
\definecolor{rageA}{HTML}{8B5CB7} 
\definecolor{rageG}{HTML}{269D90} 
\definecolor{rageE}{HTML}{D68A35} 

\newcommand{\RAGE}{%
  \textcolor{rageR}{R}%
  \textcolor{rageA}{A}%
  \textcolor{rageG}{G}%
  \textcolor{rageE}{E}%
}

\title{
\RAGE:
\textcolor{rageR}{R}esidual-%
\textcolor{rageA}{A}daptive
\textcolor{rageG}{G}ating for
Flow-Based Image
\textcolor{rageE}{E}diting
}

\author[]{Ruofeng Mei}
\author[*]{Xiangyu Xu}

\affiliation[]{Xi'an Jiaotong University}

\contribution[*]{Corresponding author}

\newcommand{\modelname}{RAGE}

\abstract{
\vspace{-5pt}
Recent inversion-free, flow-based image editing methods evolve the
source image toward the target via a differential velocity field.
We observe that this velocity field does not explicitly account
for which regions should be preserved, leading to unintended drift
in non-editing regions. To address this, we formulate source fidelity
control as a stochastic optimal control problem in which a selective
fidelity term on the editing residual is enforced along the editing
trajectory. Solving this problem and approximating its solution
yields the Residual Gating Principle: each latent component is
pulled back toward the source with a gain that decays exponentially
with the local residual magnitude, so that small deviations are
suppressed while large, semantically meaningful edits pass through
unimpeded. The proposed method, \modelname{}, requires only one
lightweight gated interpolation per sampling step, involves no mask
estimation, attention maps, retraining, or architecture-specific
modifications, and is applied on top of FlowEdit-based editors.
On PIE-Bench, \modelname{} achieves the best overall rank among 17
methods and substantially improves background preservation on both
Stable Diffusion~3 and FLUX backbones while maintaining competitive
editing quality.
\vspace{-8pt}
}

\usepackage[T1]{fontenc}
\usepackage{url}
\usepackage{graphicx}
\usepackage{xcolor}
\usepackage{amsmath}
\usepackage{amsfonts}
\usepackage{amssymb}
\usepackage{amsthm}
\usepackage{booktabs}
\usepackage{multirow}
\usepackage{array}
\usepackage{placeins}

\theoremstyle{definition}
\newtheorem{lemma}{Lemma}
\newcounter{algorithm}
\renewcommand{\thealgorithm}{\arabic{algorithm}}

\begin{document}

\maketitle

\section{Introduction}\label{sec:introduction}

Recently, inversion-free flow-based image editing
methods~\citep{kulikov2025flowedit,kim2025flowalign} have enabled
flexible editing guided by text prompts. They directly solve an
ordinary differential equation (ODE) from a source image to a target
image, without requiring an explicit inversion step. Despite their
different formulations, these methods share a common underlying
structure: the editing process is driven by a differential velocity
field $V_\theta$, which characterizes the difference between the
generative directions conditioned on the target and on the source.

However, flow matching learns transport at the distribution level and
does not explicitly distinguish which regions of a single given source
image should be modified and which should remain unchanged. As a
result, unintended drift occurs in regions unrelated to the edit.
Since each editing step depends on the current state, this drift also
accumulates over the editing trajectory. Even when an edit is
semantically localized, backgrounds may shift, textures may change,
and fine details may be altered (see Fig.~\ref{fig:motivation}).

Existing methods mitigate this drift from different perspectives.
FlowEdit~\citep{kulikov2025flowedit} employs different classifier-free
guidance scales for the source and target and truncates the sampling
timesteps, while other methods rely on attention
injection~\citep{hertz2023prompt,tumanyan2023plug} or explicit mask
estimation~\citep{couairon2023diffedit}.
FlowAlign~\citep{kim2025flowalign} addresses this drift through an
optimal control formulation: it encourages the velocity to remain
close to that of the base editor and imposes a quadratic penalty on
the final editing residual. Here, the editing residual refers
to the deviation of the editing state from the source image.

This terminal penalty encourages the entire final result to remain
close to the source image. In regions that must change to accomplish
the edit, this creates a conflict: deviations required by the edit are
penalized alongside unintended drift.

Furthermore, the penalty acts only on the final result, whereas drift
accumulates progressively during the editing process. When drift is
still small, it can be distinguished from intended edits. As it
accumulates, however, this distinction gradually fades, and the drift
becomes increasingly difficult to correct. Yet the terminal penalty
addresses drift only at the end of the editing process. Therefore,
fidelity preservation should be selective and should operate
throughout the editing trajectory, suppressing drift as soon as it
arises.

To address these challenges, we propose \textbf{\modelname{}}, a \textbf{r}esidual-\textbf{a}daptive \textbf{g}ating framework for flow-based image \textbf{e}diting that enables selective fidelity preservation throughout the editing trajectory.
We observe that the editing residual of the current state provides a
signal for distinguishing intended edits from unintended drift. The first
modification uses this signal to make fidelity preservation selective.
We replace the quadratic terminal cost with a fidelity term defined on
the editing residual. This term pulls latent components with small
residuals back toward the source image while leaving those with large
residuals unimpeded. The second modification addresses the limitation
of constraining only the final result. We replace the fixed terminal
time with a random terminal time, allowing fidelity preservation to
operate throughout the editing trajectory and thereby yielding a
stochastic optimal control problem.

Solving this stochastic optimal control problem and approximating its
solution yields a simple form that we call the \textbf{Residual Gating
Principle}: each latent component is pulled back toward the source
image with a gain that decays exponentially as its residual magnitude
increases. This suppresses small unintended drift while preserving
large intended edits. The resulting algorithm requires only one
additional gated interpolation per sampling step, without mask
estimation, attention maps, retraining, or changes to the network
architecture. It is applied on top of the base editor's update at each
step. We implement \modelname{} on FlowEdit~\citep{kulikov2025flowedit}
with both FLUX and Stable Diffusion~3 (SD3) backbones and achieve the
best overall rank on PIE-Bench.

Our contributions are summarized as follows:
\begin{itemize}
  \item We present a unified view of inversion-free flow-based
    editing, analyze the limitation of the quadratic terminal penalty
    used in FlowAlign, and identify the per-component residual
    magnitude as an intrinsic signal for distinguishing intended edits
    from unintended drift.
  \item Building upon the control formulation of FlowAlign, we
    formulate a stochastic optimal control problem with a selective
    fidelity term and a random terminal time, enabling fidelity
    preservation throughout the editing trajectory. The resulting
    algorithm requires only one lightweight gated interpolation at
    each sampling step.
  \item \modelname{} achieves the best overall rank among 17 methods
    on PIE-Bench. Across both SD3 and FLUX backbones, it delivers
    substantial and consistent improvements in background
    preservation over the base editor while maintaining competitive
    editing quality.
\end{itemize}

\section{Related Work}

\subsection{Text-Guided Image Editing with Diffusion and Flow
Models}
Text-guided image editing with pretrained generative models has been
extensively studied. Early diffusion-based approaches manipulate the
internal computation of the denoising network:
Prompt-to-Prompt~\citep{hertz2023prompt} and
Plug-and-Play~\citep{tumanyan2023plug} inject or replace attention
maps and intermediate features to constrain the edit, while
MasaCtrl~\citep{cao2023masactrl} introduces mutual self-attention for
non-rigid editing. DiffEdit~\citep{couairon2023diffedit} instead
estimates an explicit spatial mask to restrict where edits may
occur. Instruction-based methods such as
InstructPix2Pix~\citep{brooks2023instpix2pix} finetune the
generative model on paired editing data. A complementary line of work improves the inversion
process itself: Direct Inversion~\citep{ju2024direct} corrects the
source branch of inversion-based pipelines with minimal code
changes.

\subsection{Inversion-Based and Inversion-Free Methods}
With the emergence of flow matching~\citep{lipman2023flow} as the
backbone of state-of-the-art text-to-image models, editing methods
have been adapted to rectified-flow dynamics. Inversion-based flow
methods first map the source image to a noise representation and
then regenerate under the target prompt:
RF-Inversion~\citep{rout2024rfinversion} performs rectified-flow
inversion with dynamic optimal control derived via a linear quadratic
regulator, while RFEdit~\citep{wang2024rfedit} and
FireFlow~\citep{deng2024fireflow} improve the accuracy or efficiency of
reverse-ODE inversion;
FTEdit~\citep{xu2025ftedit} introduces a two-stage inversion with
invariance control for flow transformers; and
DNAEdit~\citep{xie2025dnaedit} replaces simulation-based inversion
by directly refining the desired Gaussian noise in the noise domain.
Inversion-free methods bypass the source-to-noise mapping entirely:
FlowEdit~\citep{kulikov2025flowedit} constructs a direct ODE between
source and target distributions;
FlowAlign~\citep{kim2025flowalign} adds source-consistency
regularization at the trajectory's terminal point; and
InfEdit~\citep{xu2024infedit} performs inversion-free editing with
consistency models. More recently,
SAM-Flow~\citep{cui2026samflow} restricts the differential velocity
to attention-derived soft masks and anchors the remaining regions
to the source latent, obtaining the masks from an auxiliary
editing pass. Among these methods, FlowAlign is the closest to
ours: it treats source consistency as part of the editing dynamics,
through a terminal penalty in an optimal control formulation.
\modelname{} keeps this control formulation but replaces the
quadratic terminal penalty with a selective fidelity term on the
editing residual and a random terminal time. Fidelity is thus
enforced along the trajectory, and preservation is gated by the
residual rather than uniform. Because the residual alone decides what
to preserve, \modelname{} requires no masks or attention maps. It is applied on top of FlowEdit-based editors.

\section{Method}

\begin{figure}[t]
  \centering
  \includegraphics[width=0.8\linewidth]{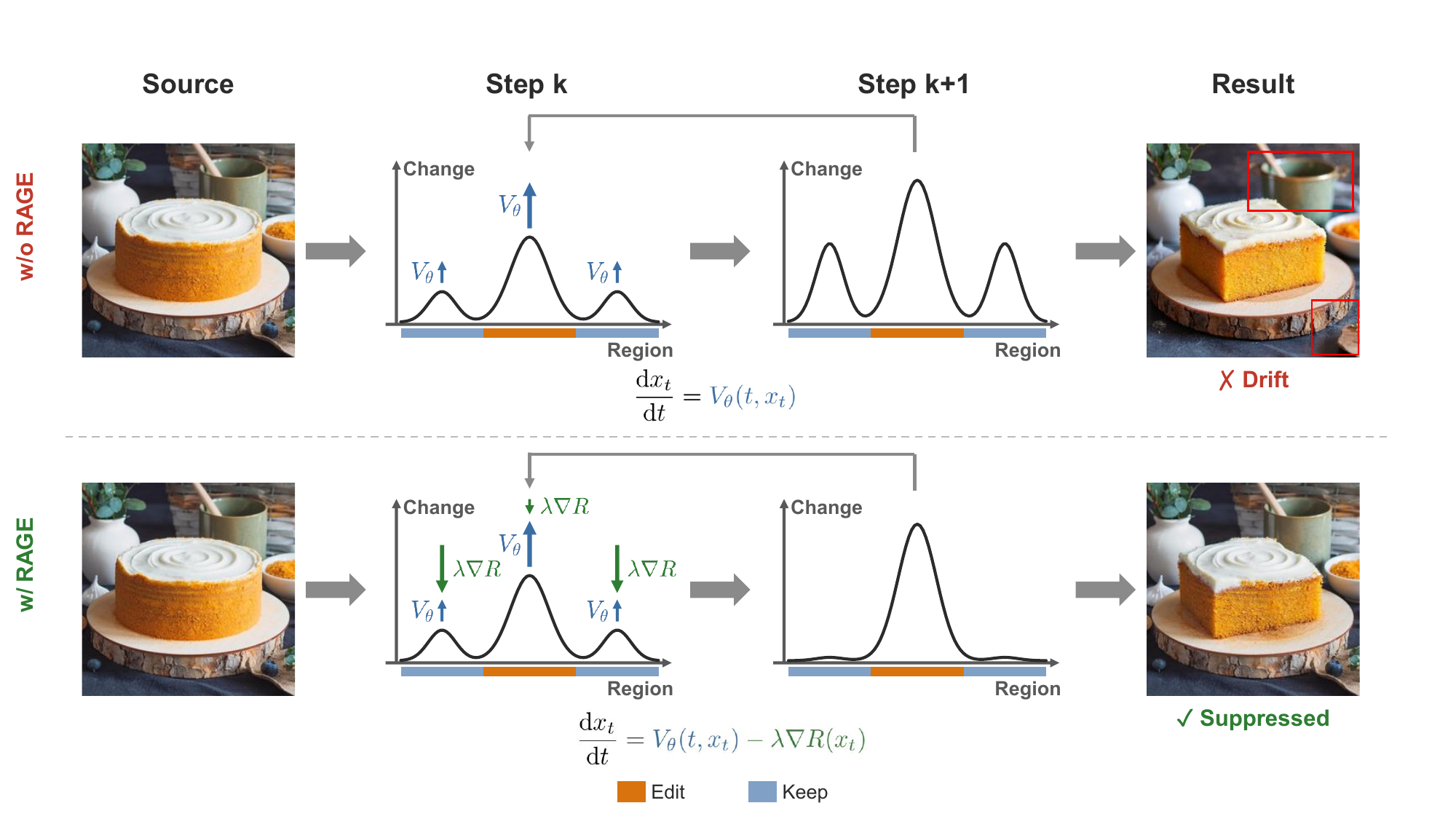}
  \caption{Without \modelname{} (top), the base editing velocity
    $V_\theta$ causes unintended drift in non-editing regions that
    accumulates over timesteps. With \modelname{} (bottom), the
    source-aware correction $\lambda\nabla R$ selectively suppresses
    small deviations while leaving large, meaningful edits unimpeded.}
  \label{fig:motivation}
\end{figure}

\subsection{Preliminaries: Flow Matching}
 
Flow Matching~\citep{lipman2023flow} constructs a continuous transport
between a simple prior distribution $p_0(x_0)$ (typically Gaussian)
and the data distribution $p_1(x_1)$, via a time-dependent velocity
field. Given $x_0 \sim p_0$ and $x_1 \sim p_1$, the interpolant
$x_t = t x_1 + (1-t) x_0$ induces the marginal velocity field
$u^{\mathrm{gt}}(t,x) := \mathbb{E}[x_1 - x_0 \mid x_t = x]$.
A neural network $u_\theta(t, x_t)$ is trained to approximate this
field by minimizing
\[
  L_{\mathrm{CFM}}
  = \mathbb{E}\,\bigl\|u_\theta(t,x_t) - (x_1 - x_0)\bigr\|_2^2.
\]
Once trained, sampling proceeds by solving the forward ODE
$\mathrm{d}x_t / \mathrm{d}t = u_\theta(t,x_t)$ from $x_0 \sim p_0$ at $t=0$
to obtain $x_1 \sim p_1$ at $t=1$. In the context of text-to-image generation,
the velocity field is additionally conditioned on a text prompt $c$,
yielding $u_\theta(t, x_t \mid c)$, which enables sampling from the
conditional distribution $p_1(x_1 \mid c)$.

\subsection{A Unified View of Flow-Based Editing}

\subsubsection{Problem Setup}

Text-guided image editing takes as input a source image
$x_{\mathrm{src}}$, a source prompt $c_{\mathrm{src}}$ describing
its original content, and a target prompt $c_{\mathrm{tar}}$
specifying the desired edit. The goal is to generate an edited image
$x_{\mathrm{tar}}$ that conforms to $c_{\mathrm{tar}}$ while
preserving the content of $x_{\mathrm{src}}$ that is unrelated
to the edit. Since the flow models considered here operate in
the latent space of a variational autoencoder (VAE), we formulate
the editing process in this space. Throughout the following
discussion, $x_{\mathrm{src}}$ and $x_{\mathrm{tar}}$ denote
the corresponding VAE latents.

\subsubsection{From Inversion-Based to Inversion-Free Methods}

A common approach to flow-based image editing is to invert the
source image into noise and then generate an image conditioned
on the target prompt from the recovered noise. Inversion-based
methods implement this in two stages. They first invert the
source image to a noise representation
$\varepsilon_{\mathrm{inv}}$ by solving the reverse ODE
$\mathrm{d}q_t/\mathrm{d}t
= u_\theta(t, q_t \mid c_{\mathrm{src}})$
from $q_1 = x_{\mathrm{src}}$ at $t = 1$ to $t = 0$.
They then solve the forward ODE
$\mathrm{d}p_t/\mathrm{d}t
= u_\theta(t, p_t \mid c_{\mathrm{tar}})$
from $t = 0$ to $t = 1$, starting from the same noise
$q_0 = p_0 = \varepsilon_{\mathrm{inv}}$ to obtain
the edited image $p_1$.

As shown by FlowEdit~\citep{kulikov2025flowedit}, this two-stage
process can be reformulated as a single ODE that directly evolves
from the source image to the target. Defining
$x_t = p_t - q_t + x_{\mathrm{src}}$ gives
$x_0 = x_{\mathrm{src}}$ and $x_1 = p_1$.
Subtracting the two ODEs yields
\begin{equation}\label{eq:edit_ode}
\begin{aligned}
  \frac{\mathrm{d}x_t}{\mathrm{d}t}
  ={} & u_\theta\!\bigl(t,\;x_t + q_t
      - x_{\mathrm{src}} \mid c_{\mathrm{tar}}\bigr) \\
  & - u_\theta\!\bigl(t,\;q_t \mid c_{\mathrm{src}}\bigr),
\end{aligned}
\end{equation}
where the right-hand side defines a differential velocity
field, given by the difference between the target-conditioned
and source-conditioned velocities. However, evaluating
\eqref{eq:edit_ode} still requires the inversion trajectory
$q_t$, so the overall process involves two ODE solves.

Inversion-free methods, such as FlowEdit~\citep{kulikov2025flowedit}
and FlowAlign~\citep{kim2025flowalign}, avoid explicit inversion
by setting $q_t = t\,x_{\mathrm{src}} + (1-t)\,\varepsilon$
with $\varepsilon \sim \mathcal{N}(0,I)$, instead of solving the
inversion ODE, and accordingly $p_t = x_t - x_{\mathrm{src}} + q_t$.
This leaves a single ODE solve for editing. While these methods differ
in their guidance and regularization strategies, their
editing dynamics can be described by an expected velocity field:
\begin{equation}\label{eq:unified}
\begin{aligned}
  \frac{\mathrm{d}x_t}{\mathrm{d}t}
  &= V_\theta(t,x_t), \qquad x_0 = x_{\mathrm{src}}, \\
  V_\theta(t,x_t)
  &= \mathbb{E}_{\varepsilon\sim\mathcal{N}(0,I)}
  \left[v_\theta^{\mathrm{edit}}(t,x_t;\varepsilon)\right],
\end{aligned}
\end{equation}
where $v_\theta^{\mathrm{edit}}$ denotes the velocity
computed from a single noise sample. It is based on the differential
velocity field in~\eqref{eq:edit_ode}, evaluated using
$q_t$, with any method-specific guidance or
regularization. In practice, the expectation is approximated
using one or more noise samples at each timestep.

\subsubsection{FlowAlign: Source Consistency via Optimal Control}
\label{sec:flowalign}

Although inversion-free methods avoid the cost of inversion,
the stochastic source trajectory can lead to poor source
consistency. FlowAlign~\citep{kim2025flowalign} addresses this
problem by introducing a terminal penalty through an optimal
control formulation:
\begin{equation}\label{eq:flowalign_oc}
\begin{aligned}
  \min_{u}\;\; &
  \int_0^1 \frac{1}{2}\|u_t-a_t\|_2^2\,\mathrm{d}t
  + \frac{\eta}{2}\|x_1-x_{\mathrm{src}}\|_2^2 \\
  \text{s.t.}\;\; &
  \frac{\mathrm{d}x_t}{\mathrm{d}t}=u_t,
  \qquad x_0=x_{\mathrm{src}},
\end{aligned}
\end{equation}
where $a_t := V_\theta(t,x_t)$ is the reference
velocity given by the base editor. The running cost
keeps the controlled velocity close to $a_t$, while the
terminal cost, weighted by $\eta>0$, penalizes deviations
of the final result from the source image.

Using a first-order approximation of the remaining trajectory,
FlowAlign obtains a source-consistency correction based on
denoising estimates. Its practical update takes the form
\begin{equation}\label{eq:flowalign_update}
  x^{k+1}=x^k+\tau\,\hat a_{t_k}
  -\zeta\bigl(
    \mathbb{E}[p_1\mid p_{t_k}]
    -\mathbb{E}[q_1\mid q_{t_k}]
  \bigr),
\end{equation}
where $\tau=t_{k+1}-t_k>0$ is the step size,
$\hat a_{t_k}$ is the sampled estimate of the reference
velocity, and $\zeta>0$ controls source consistency and
is held constant in practice. The two conditional
expectations are denoising estimates for the target and
source branches. Their difference satisfies
$\mathbb{E}[p_1\mid p_{t_k}]
-\mathbb{E}[q_1\mid q_{t_k}]
=x^k-x_{\mathrm{src}}+(1-t_k)\hat a_{t_k}$,
so the correction combines an adjustment to the editing
step size with a global pull toward the source.

\subsubsection{Limitation of the Quadratic Terminal Penalty}
\label{sec:limitation}
\label{sec:trajectorywide}

FlowAlign's terminal cost in~\eqref{eq:flowalign_oc} penalizes the
editing residual $e_1$, where $e_t := x_t - x_{\mathrm{src}}$
denotes the deviation of the editing state from the source latent.
This terminal penalty encourages the final edited result to remain
close to the source image as a whole. This creates a conflict in
regions that need to change: deviations required by the edit are
penalized alongside unintended changes. Since the penalty acts
uniformly and indiscriminately on all components, its correction is
not directed at where drift actually occurs. Indeed, as shown in
Sec.~\ref{sec:ablation_retreat}, during the early and middle stages of
editing, FlowAlign's source-consistency correction behaves
predominantly as a global step-size rescaling rather than a spatially
selective constraint.

Moreover, the penalty acts only on the final result. Since each
editing step depends on the current state, unintended drift
accumulates along the trajectory (Fig.~\ref{fig:motivation}). While
drift is still small, its residual magnitude can distinguish it from
intended edits; as drift accumulates, this distinction gradually
fades, making drift increasingly difficult to correct. A terminal
penalty, however, only addresses drift at the end of editing;
FlowAlign's per-step correction is derived from this penalty and
likewise targets the final result. Fidelity should therefore be
selective and enforced throughout the editing process, so that drift
is suppressed as it arises.

\subsection{Source-Aware Editing Velocity from
Stochastic Optimal Control}
\label{sec:rage}

This limitation motivates two changes to the control formulation used
by FlowAlign.

\subsubsection{A Selective Fidelity Term on the Editing Residual}
\label{sec:saturating}

The first change makes fidelity selective, so that residuals are no
longer penalized indiscriminately. In fact, residual magnitude itself
provides a signal for identifying regions that should be edited: in
regions that the edit must change, $V_\theta$ produces sustained,
large-magnitude updates, whereas in regions irrelevant to the edit,
its updates remain small. Hence, at the component level (where a
component is a scalar element of the latent), large residuals
correspond to clear editing intent, while small residuals are more
likely to originate from drift. An ideal correction mechanism should
therefore act selectively: pull small-residual components back toward
the source while leaving large-residual components unimpeded.

We express this selectivity with a fidelity term defined on the residual
and use it in place of the quadratic terminal cost of~\eqref{eq:flowalign_oc}:
\begin{equation}\label{eq:residual_penalty}
  R(x)
  = \sum_i \varphi\!\bigl(\,\bigl|x_i - x_{\mathrm{src},i}\bigr|\,\bigr),
\end{equation}
where the sum runs over all latent components,
$e_i = x_i - x_{\mathrm{src},i}$ is the $i$th component of the
residual, and $\varphi$ acts on its magnitude. We require $R$ to
satisfy two properties, corresponding respectively to the two
behaviors above:
\begin{itemize}
  \item[\textbf{(P1)}] \emph{Approximately linear pull-back at small
    residuals:} there exists a scale $\beta^{-1} > 0$ such that
    $\nabla_i R(x) \approx e_i$ whenever $|e_i| \ll \beta^{-1}$,
    i.e., $R$ is locally consistent with the quadratic fidelity
    $\tfrac{1}{2}\|x - x_{\mathrm{src}}\|_2^2$.
  \item[\textbf{(P2)}] \emph{Vanishing gradient at large residuals:}
    $\nabla_i R(x) \to 0$ as $|e_i| \to \infty$, so that decisive
    edits are not opposed.
\end{itemize}
Properties (P1) and (P2) follow the principle behind
redescending M-estimators in robust
statistics~\citep{beaton1974fitting,black1996unification}.
Both hold when the gradient of $R$ weights each residual component
by a factor that equals one at zero and decays faster than the
residual grows. We adopt the natural such factor, an exponential that decays on the
scale $\beta^{-1}$ of (P1):
\begin{equation}\label{eq:phi_practical_grad}
  \nabla_i R(x) = e^{-\beta |e_i|}\, e_i .
\end{equation}

\subsubsection{A Stochastic Optimal Control Formulation}
\label{sec:soc}

The second change addresses the limitation of constraining only the
final result: we replace the fixed terminal time in~\eqref{eq:flowalign_oc}
with a random terminal time $S$. Editing can be stopped at any time and evaluated
using the fidelity term, so the trajectory must maintain fidelity at all
times.
The resulting problem is
\begin{equation}\label{eq:soc}
\begin{aligned}
  \min_{u}\;\;
  & \mathbb{E}\!\left[
      \int_{0}^{S} \frac{1}{2}
      \bigl\|u_t - a_t\bigr\|_2^2\,\mathrm{d}t
      \;+\; \lambda\,R\bigl(x_S\bigr)
    \right] \\
  \text{s.t.}\;\;
  & \frac{\mathrm{d}x_t}{\mathrm{d}t} = u_t, \qquad
   x_0 = x_{\mathrm{src}},
\end{aligned}
\end{equation}
where $S$ follows an exponential distribution with rate $\rho$ truncated to $[0, 1]$.
The dynamics themselves are deterministic;
the only source of randomness in~\eqref{eq:soc} is the terminal time $S$. 
FlowAlign's objective is recovered by fixing the terminal time at $S = 1$
and choosing $R(x) = \tfrac{1}{2}\|x - x_{\mathrm{src}}\|_2^2$: under
$\eta = \lambda$, problem~\eqref{eq:soc} is precisely~\eqref{eq:flowalign_oc}.

Taking the expectation over $S$ converts the terminal fidelity cost
into a running penalty over the entire trajectory, whose weight is
largest at the start, where drift has the greatest impact, and decays
at rate $\rho$. We define the source-aware velocity field as the
optimal control of~\eqref{eq:soc}.

\subsubsection{Optimal Control at Stationarity}
\label{sec:stationarity}

To solve~\eqref{eq:soc}, we follow the treatment in
FlowAlign~\citep{kim2025flowalign}: we treat the reference velocity
$a_t = V_\theta(t, x_t)$ as given and characterize optimality in the sense
of first-order stationarity.
The following lemma gives the resulting optimal control:
the optimal velocity equals the reference velocity
minus a correction $q_t$, which is an exponentially weighted
average of the fidelity gradients over a future window. The
control anticipates where drift would otherwise
accumulate and counteracts it in proportion.

\begin{lemma}[Optimal control at first-order stationarity]
\label{lem:optimal_control}
Let $a_t$ be a given reference velocity process and let $S$ be the random terminal time defined in~\eqref{eq:soc}. For
problem~\eqref{eq:soc}, the control satisfying the Pontryagin
first-order stationarity conditions is, for $t \in [0, 1)$,
\begin{equation}\label{eq:optimal_control}
  u^{\star}_t = a_t - q_t,
\end{equation}
where
\begin{equation}\label{eq:qs}
\begin{aligned}
  q_t
  &= \frac{\rho\lambda
      \displaystyle\int_{t}^{1}
        e^{-\rho(\sigma - t)}\,
        \psi\bigl(x_\sigma\bigr)\,\mathrm{d}\sigma}
    {1 - e^{-\rho(1 - t)}}, \\
  \psi(x) &:= \nabla R(x).
\end{aligned}
\end{equation}
\end{lemma}

The proof is given in the Appendix.

\subsubsection{Localization: From Optimal Control to State Feedback}
\label{sec:localization}

The control in Lemma~\ref{lem:optimal_control} depends on fidelity
gradients along the future trajectory and is thus not directly
implementable. The exponential weight in~\eqref{eq:qs}, however,
concentrates near the current time. The following lemma quantifies
this: $q_t$ is approximated by the instantaneous gradient
$\lambda\nabla R(x_t)$ with an error that vanishes as
$O(1/\rho)$.

\begin{lemma}[Localization]
\label{lem:localization}
Let $q_t$ be as in Lemma~\ref{lem:optimal_control}, let
$g(t) := \psi(x_t)$, and let
$M_t := \sup_{\sigma \in [t,1]} \|\dot g(\sigma)\|$ (assumed
finite). Then
\begin{equation}
  \bigl\|q_t - \lambda\,g(t)\bigr\| \;\le\; \frac{\lambda M_t}{\rho},
\end{equation}
and consequently
\begin{equation}\label{eq:state_feedback}
\begin{aligned}
  u^{\star}_t
  &= a_t - \lambda\,\nabla R\bigl(x_t\bigr) + \varepsilon_\rho(t),
  \\
  \|\varepsilon_\rho(t)\| &\le \frac{\lambda M_t}{\rho}.
\end{aligned}
\end{equation}
\end{lemma}

The proof is given in the Appendix. For sufficiently
large $\rho$, that is, a sufficiently local correction
horizon, the optimal control collapses to an implementable state
feedback: the reference velocity corrected by the current fidelity
gradient.
Accounting for the dependence of $a_t$ on the state changes this control
only by a further $O(1/\rho)$ term (see the Appendix).

Dropping the controlled $O(\lambda M_t/\rho)$ remainder
and using $a_t = V_\theta(t, x_t)$ gives the
augmented editing ODE
\begin{equation}\label{eq:problem_formulation}
\frac{\mathrm{d}x_t}{\mathrm{d}t}
= \underbrace{V_\theta(t,x_t)}_{\text{base editing velocity}}
- \underbrace{\lambda\nabla R(x_t)}_{\text{source-aware correction}}.
\end{equation}

\subsection{The Practical Algorithm}
\label{sec:discretization}

\subsubsection{Residual-Gated Dynamics}

Substituting~\eqref{eq:phi_practical_grad}
into~\eqref{eq:problem_formulation} gives the fully explicit,
residual-gated dynamics
\begin{equation}\label{eq:gated_ode}
  \frac{\mathrm{d}x_t}{\mathrm{d}t}
  = V_\theta\!\bigl(t, x_t\bigr)
  - \lambda\, e^{-\beta\,|x_t - x_{\mathrm{src}}|}
    \odot \bigl(x_t - x_{\mathrm{src}}\bigr),
\end{equation}
where the exponential and the modulus are applied elementwise and
$\odot$ denotes the Hadamard product.

This is precisely the continuous-time form of the following
principle:

\smallskip
\noindent\textbf{Residual Gating Principle.}\;
\textit{The source-aware correction acts as a nonlinear gate: each
latent component is pulled back toward the source image with a gain
that decays exponentially with the magnitude of the local editing
residual. Regions with small residuals, where editing intent is ambiguous, are strongly attracted to the source; regions with large residuals, where editing intent is clear, pass through unimpeded.}
\smallskip

\subsubsection{Splitting Scheme}

We discretize~\eqref{eq:gated_ode} with a splitting scheme
that separates the editing proposal from the source-aware
correction. With step size $\tau$ ($t$ increases by $\tau$ per
step), the base editor first advances the trajectory,
\begin{equation}\label{eq:edit_step}
  \hat{x}^k = x^k + \tau\,V_\theta(t_k, x^k),
\end{equation}
and the correction is then applied to the resulting proposal,
\begin{equation}\label{eq:practical_update}
  \begin{aligned}
    x^{k+1}
    &= \hat{x}^k
      - \alpha\, w^k \odot \bigl(\hat{x}^k - x_{\mathrm{src}}\bigr), \\
    w^k &= e^{-\beta\,|x^k - x_{\mathrm{src}}|},
    \qquad \alpha = \tau\lambda.
  \end{aligned}
\end{equation}
Note that the gate $w^k$ is computed from the residual of $x^k$ before the
editor step and then applied to the proposal $\hat x^k$.

\subsubsection{Warm-Up and Hyperparameters}\label{warm-up}

The gate distinguishes edits from drift using residual magnitude, but this
signal is not yet available at the start of editing: the trajectory begins at
$x_{\mathrm{src}}$, all residuals are close to zero, and regions that should
be edited appear indistinguishable from those that should be preserved. A
correction applied at this stage acts uniformly across all components and
cannot yet be selective.

We delay the correction while the base editor takes a few steps on its
own. During this warm-up, residuals build up where changes are
needed, providing the contrast required for selective preservation.

In the algorithm, this is implemented by the attraction threshold $t_0$: the
first $K_0$ steps ($t_k < t_0$) perform only the editor
step~\eqref{eq:edit_step}, i.e., $x^{k+1} = \hat x^k$, and the
correction~\eqref{eq:practical_update} is applied at each subsequent step.
Our experiments use $K_0 = 4$ for both backbones; Table~\ref{tab:ablation}
shows that this setting provides a good balance between source preservation and editing strength.
The base editor's guidance configuration for these initial steps is specified
in Sec.~\ref{sec:setup}. In the control formulation,
the warm-up shifts the starting time of~\eqref{eq:soc} from $0$ to $t_0$.
As shown in the Appendix for a general starting time,
the expression~\eqref{eq:qs} does not involve $t_0$: the
correction law depends only on the segment from the current time to
the terminal time, not on when the correction was switched on. The warm-up
length therefore selects when the correction acts, but does
not alter what the correction is.

The method thus exposes three interpretable hyperparameters: the
gate strength $\alpha = \tau\lambda$ (overall correction strength;
increasing $\alpha$ strengthens source preservation), the decay rate
$\beta$ (gate sharpness; $\beta^{-1}$ is the residual scale
separating drift from intended edits, cf.\ (P1) and (P2)), and the
attraction threshold $t_0$ (when the correction switches on). The
rate $\rho$ does not appear in the final algorithm: it is absorbed
by the localization limit of Lemma~\ref{lem:localization}.

\subsubsection{Instantiations on FlowEdit-Based Editors}

Because the correction~\eqref{eq:practical_update} depends only on
the current state and the source latent, it is applied on top of the base editor's update at each step. We
instantiate \modelname{} on the inversion-free FlowEdit
framework~\citep{kulikov2025flowedit}.
Algorithm~\ref{alg:rage_flux} gives the
instantiation used in our FLUX experiments, which follows the
FlowEdit skeleton directly. Algorithm~\ref{alg:rage_sd3}
gives the SD3 instantiation,
which uses the same skeleton and additionally adopts the
source-consistency correction~\eqref{eq:flowalign_update} of FlowAlign~\citep{kim2025flowalign}.
The two algorithms differ only in the base editor: FLUX is a
guidance-distilled model that takes the guidance scale as a model input,
whereas on SD3, classifier-free guidance is applied explicitly to the target
branch.
In both algorithms, \modelname{}
consists of the steps highlighted in blue, which compute the
gate from the current residual and apply the gated
interpolation~\eqref{eq:practical_update}, while all remaining lines follow the base editor's skeleton, with the configuration detailed in Sec.~\ref{sec:setup}. The correction requires no mask estimation, no
attention maps, no retraining, and no architecture-specific
modifications, and it adds negligible overhead (one pointwise
operation per sampling step). To measure this overhead, we run the SD3 instantiation on a subset of 70 images with and without the gate, keeping all other settings fixed. Editing the subset takes 212.907~s with the base editor alone and 214.863~s with \modelname{}, an increase of 1.956~s in total, or about 0.03~s per image (0.92\%). 

\begin{figure}[!t]
\centering
\refstepcounter{algorithm}
\label{alg:rage_flux}
\begin{minipage}{0.8\linewidth}
\footnotesize
\hrule height 0.6pt
\vspace{0.3em}
{\textbf{Algorithm \thealgorithm} \modelname{} on the FlowEdit
skeleton~\citep{kulikov2025flowedit} (FLUX instantiation)}
\vspace{0.3em}
\hrule height 0.6pt
\vspace{0.35em}

\noindent
\textbf{Require:}\quad
Source image $z_{\mathrm{src}}$, Pre-trained flow model $u_{\theta}$,
VAE encoder and decoder $\mathcal{E},\mathcal{D}$,
Source/Target text embeddings $c_{\mathrm{src}},c_{\mathrm{tar}}$,
guidance schedules $w^{\mathrm{src}}_t,w^{\mathrm{tar}}_t$ and drift
scale $\gamma_t$ (base editor),
gate strength $\textcolor{blue}{\alpha}$,
decay rate $\textcolor{blue}{\beta}$,
attraction threshold $\textcolor{blue}{t_0}$.

\vspace{0.3em}
\setlength{\tabcolsep}{0pt}
\renewcommand{\arraystretch}{1.08}
\begin{tabular}{@{}r@{\hspace{0.45em}}p{0.89\linewidth}@{}}
1:  & $x_{\mathrm{src}} \leftarrow \mathcal{E}(z_{\mathrm{src}})$ \\
2:  & $x_t \leftarrow x_{\mathrm{src}}$ \\
3:  & $\textbf{for } t: 0 \rightarrow 1 \textbf{ do}$ \\
4:  & $\qquad \varepsilon \sim \mathcal{N}(0,I)$ \\
5:  & $\qquad q_t \leftarrow t x_{\mathrm{src}}+(1-t)\varepsilon$ \\
6:  & $\qquad p_t \leftarrow x_t-x_{\mathrm{src}}+q_t$ \\
7:  & $\textcolor{blue}{\qquad e \leftarrow x_t-x_{\mathrm{src}}}$ \\
8:  & $\qquad v_q \leftarrow u_{\theta}(t, q_t \mid c_{\mathrm{src}}; w^{\mathrm{src}}_t)$ \\
9:  & $\qquad v_p \leftarrow u_{\theta}(t, p_t \mid c_{\mathrm{tar}}; w^{\mathrm{tar}}_t)$ \\
10: & $\qquad x_t \leftarrow x_t+\gamma_t\left(v_p-v_q\right)dt$ \\
11: & $\textcolor{blue}{\qquad \textbf{if } t \ge t_0 \textbf{ then}} $ \\
12: & $\textcolor{blue}{\qquad\qquad x_t \leftarrow x_t - \alpha\,e^{-\beta\,|e|} \odot (x_t - x_{\mathrm{src}})}$ \\
13: & $\textbf{end for}$ \\
14: & $z_{\mathrm{edit}} \leftarrow \mathcal{D}(x_t)$
\end{tabular}

\vspace{0.35em}
\hrule height 0.6pt
\end{minipage}
\end{figure}

\begin{figure}[!t]
\centering
\refstepcounter{algorithm}
\label{alg:rage_sd3}
\begin{minipage}{0.8\linewidth}
\footnotesize
\hrule height 0.6pt
\vspace{0.3em}
{\textbf{Algorithm \thealgorithm} \modelname{} (SD3 instantiation): the FlowEdit
skeleton with CFG and the source-consistency correction of
FlowAlign~\citep{kim2025flowalign} (base-editor implementation
detail)}
\vspace{0.3em}
\hrule height 0.6pt
\vspace{0.35em}

\noindent
\textbf{Require:}\quad
Source image $z_{\mathrm{src}}$, Pre-trained flow model $u_{\theta}$,
VAE encoder and decoder $\mathcal{E}, \mathcal{D}$,
Source/Target text embeddings $c_{\mathrm{src}}, c_{\mathrm{tar}}$,
CFG scale $\omega_t$,
source consistency scale $\zeta$ (from~\citealp{kim2025flowalign}),
gate strength $\textcolor{blue}{\alpha}$,
gate decay rate $\textcolor{blue}{\beta}$,
attraction threshold $\textcolor{blue}{t_0}$.

\vspace{0.3em}
\setlength{\tabcolsep}{0pt}
\renewcommand{\arraystretch}{1.08}
\begin{tabular}{@{}r@{\hspace{0.45em}}p{0.89\linewidth}@{}}
1:  & $x_{\mathrm{src}} \leftarrow \mathcal{E}(z_{\mathrm{src}})$ \\
2:  & $x_t \leftarrow x_{\mathrm{src}}$ \\
3:  & $\textbf{for } t: 0 \rightarrow 1 \textbf{ do}$ \\
4:  & $\qquad \varepsilon \sim \mathcal{N}(0,I)$ \\
5:  & $\qquad q_t \leftarrow t x_{\mathrm{src}} + (1-t)\varepsilon$ \\
6:  & $\qquad p_t \leftarrow x_t - x_{\mathrm{src}} + q_t$ \\
7:  & $\textcolor{blue}{\qquad e \leftarrow x_t - x_{\mathrm{src}}}$ \\
8:  & $\begin{aligned}[t]
        \qquad v_p := {}&u_{\theta}(t, p_t \mid c_{\mathrm{src}}) \\
        &+ \omega_t\left[
        u_{\theta}(t, p_t \mid c_{\mathrm{tar}})
        - u_{\theta}(t, p_t \mid c_{\mathrm{src}})
        \right]
        \end{aligned}$ \\
9:  & $\qquad v_q := u_{\theta}(t, q_t \mid c_{\mathrm{src}})$ \\
10: & $\begin{aligned}[t]
        \qquad \mathbb{E}[p_1|p_t] &\leftarrow p_t + (1-t)\,v_p, \\
        \qquad \mathbb{E}[q_1|q_t] &\leftarrow q_t + (1-t)\,v_q
        \end{aligned}$ \\
11: & $\begin{aligned}[t]
        \qquad x_t \leftarrow x_t
        &+ \left(v_p - v_q\right)dt \\
        &+ \zeta\left(
        \mathbb{E}[q_1|q_t]
        - \mathbb{E}[p_1|p_t]
        \right)
        \end{aligned}$ \\
12: & $\textcolor{blue}{\qquad \textbf{if } t \ge t_0 \textbf{ then}}$ \\
13: & $\textcolor{blue}{\qquad\qquad x_t \leftarrow x_t - \alpha\,e^{-\beta\,|e|} \odot (x_t - x_{\mathrm{src}})}$ \\
14: & $\textbf{end for}$ \\
15: & $z_{\mathrm{edit}} \leftarrow \mathcal{D}(x_t)$
\end{tabular}

\vspace{0.35em}
\hrule height 0.6pt
\end{minipage}
\end{figure}

\section{Experiments}

\subsection{Experimental Setup}\label{sec:setup}

\noindent\textbf{Benchmark and Metrics.}\quad
We evaluate on PIE-Bench~\citep{ju2024direct}, a standard benchmark
comprising 700 images across ten editing categories, each annotated
with source and target prompts and a foreground mask. Following the
official protocol~\citep{ju2024direct}, we measure structure
consistency via DINO-ViT self-similarity
distance~\citep{tumanyan2022self} ($\times 10^3$, $\downarrow$),
background preservation via PSNR ($\uparrow$),
LPIPS~\citep{zhang2018unreasonable} ($\times 10^3$, $\downarrow$),
MSE ($\times 10^4$, $\downarrow$), and
SSIM~\citep{wang2004image} ($\times 10^2$, $\uparrow$) computed
outside the annotated mask, and editing faithfulness via CLIP
similarity~\citep{radford2021clip} ($\times 10^2$, $\uparrow$) over
both the whole image (Whole) and the edited region (Edited). We also
report the mean rank across all metrics (Rank Avg., $\downarrow$)
following~\citet{xie2025dnaedit}.

\noindent\textbf{Baselines.}\quad
We compare against a broad set of state-of-the-art methods spanning
three categories. UNet-based methods include
MasaCtrl~\citep{cao2023masactrl},
Plug-and-Play (PnP)~\citep{tumanyan2023plug}, and their combinations
with Direct Inversion~\citep{ju2024direct} (DI+MasaCtrl, DI+PnP),
built on SD~1.4/1.5. Inversion-based flow methods include
RF-Inversion~\citep{rout2024rfinversion},
RFEdit~\citep{wang2024rfedit},
FireFlow~\citep{deng2024fireflow},
FTEdit~\citep{xu2025ftedit}, and
DNAEdit~\citep{xie2025dnaedit}.
Inversion-free methods include
InfEdit~\citep{xu2024infedit},
FlowEdit~\citep{kulikov2025flowedit},
and FlowAlign~\citep{kim2025flowalign}.
We also include
InstructPix2Pix~\citep{brooks2023instpix2pix} as a representative
instruction-based baseline. Results for
FlowEdit~(SD3, FLUX) and FlowAlign~(SD3) are reproduced using their
official implementations; results for all other baselines are taken
from~\citet{xie2025dnaedit}. We report two instantiations of
\modelname{}, both built on the inversion-free FlowEdit
framework~\citep{kulikov2025flowedit}: \modelname{}~(FLUX) on
FLUX.1-dev (Algorithm~\ref{alg:rage_flux}) and
\modelname{}~(SD3) on Stable
Diffusion~3, the latter additionally adopting the source-consistency
term of FlowAlign~\citep{kim2025flowalign}
(Algorithm~\ref{alg:rage_sd3}).

\noindent\textbf{Implementation Details.}\quad
Both instantiations of \modelname{} are implemented in the FlowEdit
codebase~\citep{kulikov2025flowedit}. \modelname{}~(FLUX) applies the
residual-gated correction~\eqref{eq:practical_update} on top of
FlowEdit with FLUX.1-dev as the backbone
(Algorithm~\ref{alg:rage_flux}).
\modelname{}~(SD3) uses Stable Diffusion~3 (medium) as the backbone
with the FlowEdit skeleton and additionally adopts the
source-consistency correction of
FlowAlign~\citep{kim2025flowalign}
(Algorithm~\ref{alg:rage_sd3}).
In both cases, \modelname{} adds the highlighted correction
steps of Algorithms~\ref{alg:rage_flux}
and~\ref{alg:rage_sd3}. 

We adopt the default FlowEdit configuration for FLUX
(28 steps, 24 active,
$(w^{\mathrm{src}}_t, w^{\mathrm{tar}}_t) = (1.5,\;5.5)$,
$\gamma_t = 1$) except during the warm-up phase ($K_0 = 4$ steps),
where we set both guidance scales to $1.0$ and $\gamma_t = 3.2$.
For SD3 we follow the FlowAlign defaults
(33 steps, $\zeta = 0.01$; \citealp{kim2025flowalign}) but set the
CFG scale to $\omega_t = 10$ during warm-up ($K_0 = 4$ steps)
and $\omega_t = 20$ afterwards.

The gate strength $\alpha$ is set to $1.0$ for both backbones. For
SD3, the decay rate is fixed to $\beta = 100$. For FLUX, $\beta$ is
adapted per step as
$\beta = 3 / \overline{|\tau V_\theta|}_{0.9}$, where $\overline{|\tau V_\theta|}_{0.9}$
denotes the mean of the top-$10\%$ absolute values of the current
velocity step; this normalization ensures that the gate threshold
scales with the typical edit magnitude, accommodating the different
latent-space statistics of the two backbones. No per-image or
per-category tuning is performed. The correction is applied
element-wise in the VAE latent space and adds negligible overhead
(a single pointwise operation per step). All experiments are
conducted on a single GPU with 48\,GB memory.

\subsection{Quantitative Comparison}

Table~\ref{tab:quantitative_comparison} summarizes the results on
PIE-Bench. \modelname{}~(SD3) achieves the best Rank Avg.\ among all
17 methods, and \modelname{}~(FLUX) ties for second.

The main advantage lies in source preservation. The two
instantiations occupy the top two positions on Structure Distance and
on all four background preservation metrics. Compared with
FlowEdit~\citep{kulikov2025flowedit}, the base editor on the same
backbone, \modelname{} reduces both LPIPS and MSE to about one third
of FlowEdit's values on FLUX and SD3, achieving a better balance
between source preservation and editing strength. Both
instantiations also substantially outperform FlowEdit in Rank Avg.
These improvements come from a single lightweight residual-gated
correction, without mask estimation or attention maps.

Notably, FlowAlign~\citep{kim2025flowalign} also explicitly targets
source consistency, but enforces it only at the terminal point.
\modelname{}~(SD3) outperforms FlowAlign on all seven metrics,
including both CLIP scores.

FlowEdit~(SD3) achieves the highest CLIP scores, but it propagates
the target prompt across the entire image, which noticeably weakens
background preservation. In contrast, \modelname{}~(SD3) ranks second
among all methods in CLIP Whole while leading on the preservation
metrics.

\begin{table*}[!t]
\centering
\footnotesize
\caption{Quantitative comparison on PIE-Bench~\citep{ju2024direct}.
  \textbf{Bold}: best. \underline{Underline}: second best.
  Structure Distance ($\times 10^3$), LPIPS ($\times 10^3$),
  MSE ($\times 10^4$), SSIM ($\times 10^2$),
  CLIP Sim.\ ($\times 10^2$).
  Background preservation metrics are computed over the region outside
  the annotated mask.
  Rank Avg.\ is computed by globally ranking all methods on each of
  the seven individual metrics and averaging the resulting ranks with
  equal weights. Rankings are computed using the original unrounded
  metric values.
  Results for FlowEdit~(SD3, FLUX) and FlowAlign~(SD3) are
  reproduced using their official implementations; results for all
  other baselines are taken from~\citet{xie2025dnaedit}.}
\label{tab:quantitative_comparison}
\setlength{\tabcolsep}{2.5pt}
\renewcommand{\arraystretch}{1.03}
\resizebox{\textwidth}{!}{%
\begin{tabular}{@{}llcccccccr@{}}
\toprule
\multirow{2}{*}{Method} & \multirow{2}{*}{Model}
  & Struct. & \multicolumn{4}{c}{Background Preservation}
  & \multicolumn{2}{c}{CLIP Sim.}
  & \multirow{2}{*}{Rank Avg.$\downarrow$} \\
\cmidrule(lr){3-3}
\cmidrule(lr){4-7}
\cmidrule(lr){8-9}
& & Dist.$\downarrow$
  & PSNR$\uparrow$
  & LPIPS$\downarrow$
  & MSE$\downarrow$
  & SSIM$\uparrow$
  & Whole$\uparrow$
  & Edited$\uparrow$
  & \\
\midrule

\modelname{} & SD3
  & \underline{10.69}
  & \textbf{29.84}
  & \underline{32.85}
  & \textbf{24.22}
  & \underline{92.12}
  & \underline{26.48}
  & 22.76
  & \textbf{2.14} \\

\modelname{} & FLUX
  & \textbf{8.80}
  & \underline{28.28}
  & \textbf{32.17}
  & \underline{29.30}
  & \textbf{93.27}
  & 25.28
  & 21.87
  & \underline{4.71} \\

\midrule

DNAEdit~\citep{xie2025dnaedit} & SD3.5
  & 14.19
  & 26.66
  & 74.57
  & 32.76
  & 88.63
  & 25.63
  & 22.71
  & \underline{4.71} \\

FTEdit~\citep{xu2025ftedit} & SD3.5
  & 18.17
  & 26.62
  & 80.55
  & 40.24
  & 91.50
  & 25.74
  & 22.27
  & 5.57 \\

DNAEdit~\citep{xie2025dnaedit} & FLUX
  & 18.87
  & 24.99
  & 95.06
  & 50.45
  & 85.71
  & 25.79
  & 22.87
  & 6.14 \\

FlowAlign~\citep{kim2025flowalign} & SD3
  & 28.95
  & 25.33
  & 46.53
  & 42.25
  & 88.30
  & 26.28
  & 22.36
  & 6.43 \\

InfEdit~\citep{xu2024infedit} & LCM
  & 19.31
  & 27.31
  & 56.32
  & 47.80
  & 85.30
  & 24.90
  & 22.14
  & 7.43 \\

FlowEdit~\citep{kulikov2025flowedit} & SD3
  & 27.45
  & 22.18
  & 101.72
  & 87.55
  & 83.89
  & \textbf{27.48}
  & \textbf{23.04}
  & 8.57 \\

RFEdit~\citep{wang2024rfedit} & FLUX
  & 21.79
  & 24.83
  & 113.15
  & 52.46
  & 83.38
  & 25.57
  & 22.26
  & 9.57 \\

FireFlow~\citep{deng2024fireflow} & FLUX
  & 29.03
  & 23.33
  & 133.40
  & 70.83
  & 81.22
  & 26.19
  & \underline{22.99}
  & 9.57 \\

FlowEdit~\citep{kulikov2025flowedit} & FLUX
  & 28.85
  & 21.87
  & 109.80
  & 97.19
  & 83.94
  & 26.19
  & 22.88
  & 10.00 \\

DI+PnP~\citep{ju2024direct,tumanyan2023plug} & SD1.5
  & 23.35
  & 22.46
  & 105.51
  & 79.94
  & 79.88
  & 25.49
  & 22.64
  & 10.14 \\

DI+MasaCtrl~\citep{ju2024direct,cao2023masactrl} & SD1.4
  & 23.58
  & 22.68
  & 87.41
  & 80.63
  & 81.51
  & 24.39
  & 21.41
  & 11.29 \\

PnP~\citep{tumanyan2023plug} & SD1.5
  & 27.35
  & 22.31
  & 112.76
  & 82.95
  & 79.25
  & 25.41
  & 22.52
  & 11.71 \\

MasaCtrl~\citep{cao2023masactrl} & SD1.4
  & 27.12
  & 22.19
  & 105.44
  & 86.37
  & 79.91
  & 24.03
  & 21.15
  & 13.14 \\

RF-Inv~\citep{rout2024rfinversion} & FLUX
  & 42.29
  & 20.20
  & 179.54
  & 139.20
  & 69.91
  & 24.57
  & 22.20
  & 15.57 \\

InsP2P~\citep{brooks2023instpix2pix} & ---
  & 58.13
  & 20.80
  & 159.23
  & 221.30
  & 76.47
  & 23.61
  & 21.68
  & 16.29 \\

\bottomrule
\end{tabular}%
}
\end{table*}

\subsection{Qualitative Comparison}

\begin{figure*}[!tp]
  \centering
  \includegraphics[width=\textwidth]{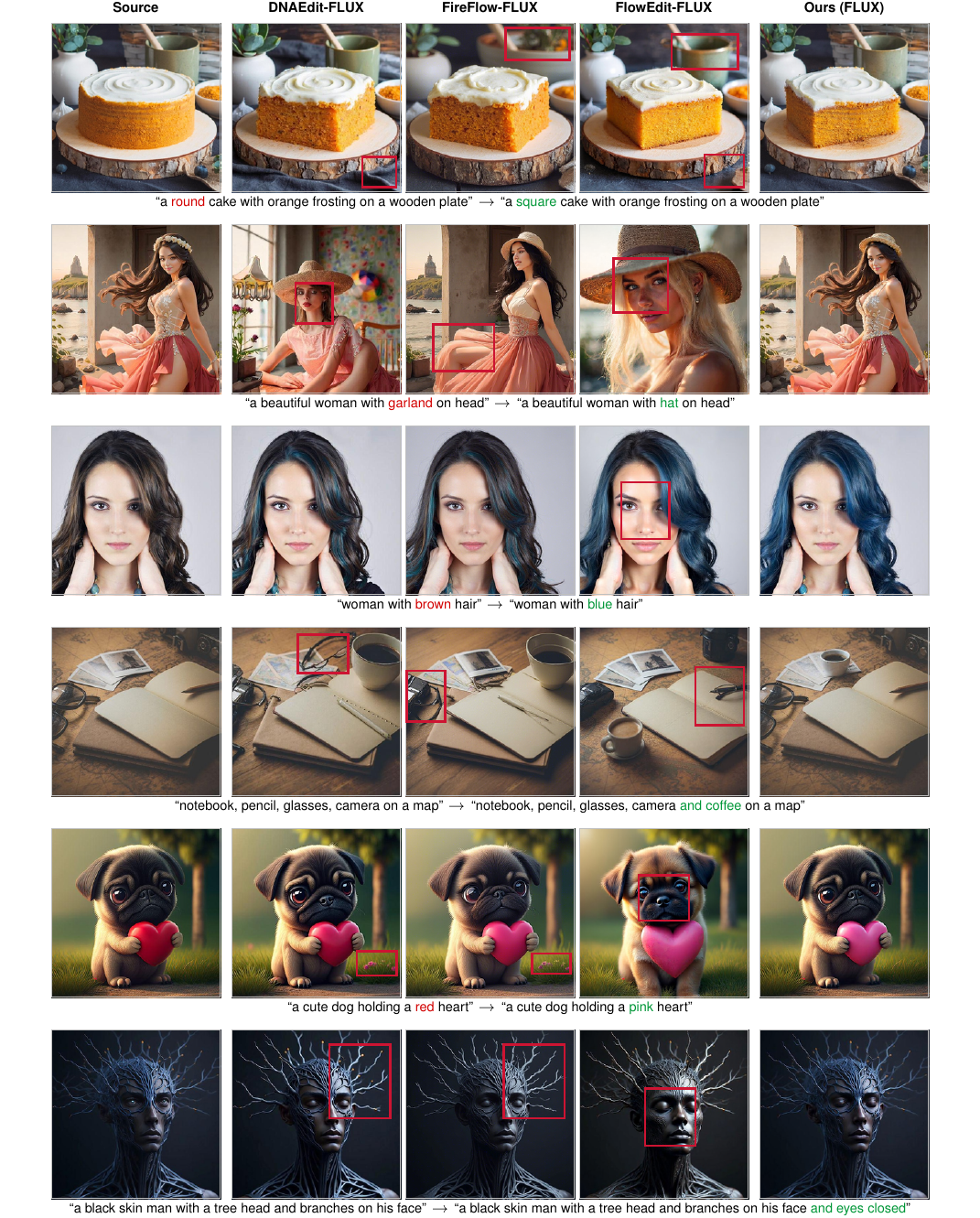}
  \caption{Qualitative comparison on PIE-Bench with FLUX.1-dev. From
    left to right: Source, DNAEdit-FLUX, FireFlow-FLUX, FlowEdit-FLUX,
    and Ours~(FLUX). Source and target prompts appear below each
    example, with the changed concepts highlighted. Red boxes mark
    unintended changes outside the edited region. The selected rows
    cover shape modification, object replacement, color editing, object
    addition, and expression editing.}
  \label{fig:qual_flux}
\end{figure*}

\begin{figure*}[!tp]
  \centering
  \includegraphics[width=\textwidth]{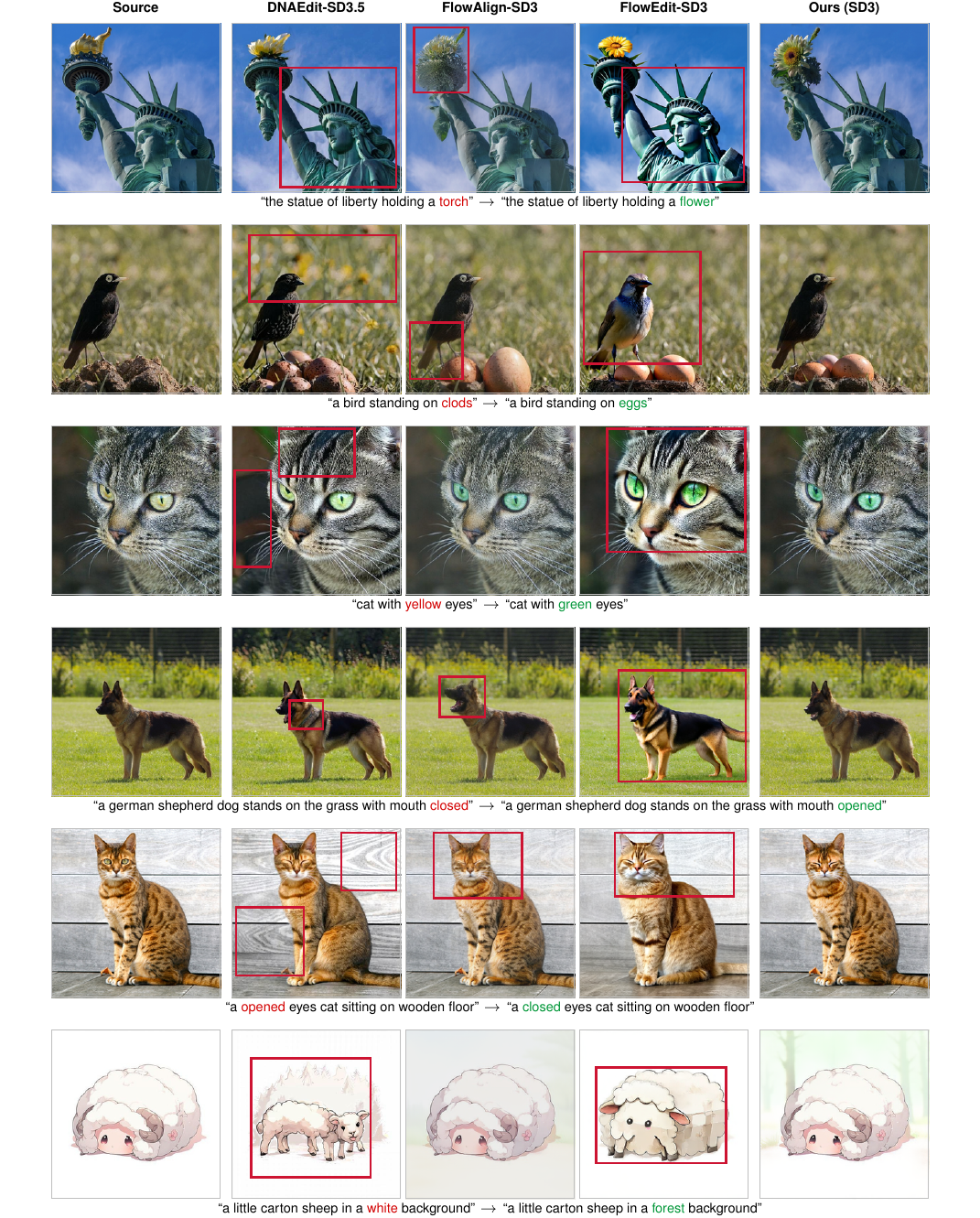}
  \caption{Qualitative comparison on PIE-Bench with Stable
    Diffusion. From left to right: Source, DNAEdit-SD3.5, FlowAlign-SD3,
    FlowEdit-SD3, and Ours~(SD3). Source and target prompts
    appear below each example, with the changed concepts highlighted.
    Red boxes mark unintended changes outside the edited region. The
    selected rows cover object replacement, color editing, expression
    editing, and background replacement.}
  \label{fig:qual_sd3}
\end{figure*}

Figs.~\ref{fig:qual_flux} and~\ref{fig:qual_sd3} present twelve
examples from PIE-Bench, one figure for each backbone (FLUX.1-dev and
Stable Diffusion), covering shape modification, object replacement,
color editing, object addition, expression editing, and background
replacement. We compare the results from two perspectives: whether
regions outside the intended edit, including the background, nearby
objects, subject identity, and lighting, remain consistent with the
source image, and whether the requested edit is fully realized. The
former reveals unintended drift, while the latter checks whether
\modelname{} suppresses this drift while still completing the edit.

The baselines exhibit such drift even when the requested edit is
minor. On FLUX (Fig.~\ref{fig:qual_flux}), when only the color of the
heart should change, FlowEdit turns the pug into a noticeably
different dog (row 5), and when the task is simply to add a cup, all
three baselines also rearrange the other objects on the desk (row 4). On SD
(Fig.~\ref{fig:qual_sd3}), when only the torch should become a
flower, FlowEdit re-renders the entire Statue of Liberty (row 1), and
when only the clods beneath the bird should become eggs, it changes
the black bird into a different species (row 2). Some baselines also
fail to complete the intended edit. In the sheep example
(Fig.~\ref{fig:qual_sd3}, row 6), the ``forest'' generated by
FlowAlign is only a faint gray-green tint with no visible trees.

In contrast, \modelname{} completes these edits while keeping the
rest of the image close to the source. On FLUX, the garland becomes a
hat while the woman's pose, dress, and the seaside background remain
nearly unchanged (row 2). The cake becomes square while the
frosting texture, the mug, and the vase stay the same (row 1). The hair turns
entirely blue, rather than showing only partial blue highlights as in
DNAEdit and FireFlow, while the face and background remain intact (row 3). A cup
of coffee is added while the rest of the desk stays as it was (row 4). This is exactly the behavior the
residual gate is designed for: regions that need editing are allowed
to change, and the rest of the image is held to the source.

The same pattern holds on SD (Fig.~\ref{fig:qual_sd3}), whether the
edit is small or large. Changing the cat's eye color affects only its
irises (row 3), whereas DNAEdit and FlowEdit also modify other parts
of the cat. In the background replacement example, the white
background is replaced with a forest while the cartoon sheep remains
intact (row 6), showing that the gate does not simply favor small
edits. When the edit target is right next to an object that must be
preserved, as with the torch held by the Statue of Liberty and the
clods beneath the bird (rows 1 and 2), \modelname{} completes the
replacement without altering the statue or the bird. In the
closed-eyes example (row 5), the expression changes while the cat's
pose, fur pattern, and the wooden wall remain unaffected.

Overall, the qualitative results show the same pattern as
Table~\ref{tab:quantitative_comparison}: on both backbones,
\modelname{} removes most of the unintended changes introduced by the
baselines while fully realizing the requested edits.

\subsection{Ablation Study}
\label{sec:ablation}

We investigate three key design choices of \modelname{} on both the SD3 and FLUX instantiations: residual-gated correction, warm-up, and fidelity throughout the trajectory. Table~\ref{tab:ablation} summarizes the ablation results.

\begin{table*}[!t]
\centering
\footnotesize
\caption{Component ablation on PIE-Bench.
  Metrics follow the conventions of
  Table~\ref{tab:quantitative_comparison}.
  All variants share the same base editor; only the specified correction, warm-up, or fidelity settings differ.
  The full model row for each backbone reproduces the corresponding
  entry from Table~\ref{tab:quantitative_comparison}.}
\label{tab:ablation}
\setlength{\tabcolsep}{5.5pt}
\renewcommand{\arraystretch}{1.05}
\begin{tabular}{@{}lccccccc@{}}
\toprule
\multirow{2}{*}{Variant}
  & Struct. & \multicolumn{4}{c}{Background Preservation}
  & \multicolumn{2}{c}{CLIP Sim.} \\
\cmidrule(lr){2-2}
\cmidrule(lr){3-6}
\cmidrule(lr){7-8}
  & Dist.$\downarrow$
  & PSNR$\uparrow$
  & LPIPS$\downarrow$
  & MSE$\downarrow$
  & SSIM$\uparrow$
  & Whole$\uparrow$
  & Edited$\uparrow$ \\
\midrule

\multicolumn{8}{@{}l}{\textit{SD3 instantiation
  (Algorithm~\ref{alg:rage_sd3})}} \\[2pt]

\modelname{} (full)
  & 10.69 & 29.84 & 32.85 & 24.22 & 92.12
  & 26.48 & 22.76 \\

w/o Correction
  & 25.79 & 23.76 & 90.05 & 67.74 & 86.49
  & 26.90 & 22.99 \\

w/o Warm-up
  & 5.89 & 32.08 & 19.88 & 12.28 & 93.48
  & 25.93 & 22.26 \\

\;\;+ original schedule
  & 5.72 & 32.20 & 19.51 & 11.94 & 93.51
  & 25.88 & 22.19 \\

Endpoint-only fidelity
  & 16.30 & 27.22 & 53.91 & 42.52 & 90.36
  & 26.62 & 22.80 \\

\midrule

\multicolumn{8}{@{}l}{\textit{FLUX instantiation
  (Algorithm~\ref{alg:rage_flux})}} \\[2pt]

\modelname{} (full)
  & 8.80 & 28.28 & 32.17 & 29.30 & 93.27
  & 25.28 & 21.87 \\

w/o Correction
  & 17.61 & 24.70 & 69.78 & 53.97 & 88.68
  & 25.90 & 22.59 \\

w/o Warm-up
  & 0.86 & 35.11 & 9.84 & 5.98 & 96.09
  & 23.56 & 20.40 \\

\;\;+ original schedule
  & 0.78 & 35.32 & 9.61 & 5.75 & 96.01
  & 23.50 & 20.35 \\

Endpoint-only fidelity
  & 9.67 & 27.53 & 38.77 & 31.73 & 92.51
  & 25.19 & 21.90 \\

\bottomrule
\end{tabular}
\end{table*}

\medskip
\noindent\textbf{Residual-gated correction.}
To evaluate the role of residual-gated correction, we consider the \emph{w/o Correction} variant, which sets $\alpha = 0$ in \eqref{eq:practical_update}, leaving only the updates from the base editor. As shown in Table~\ref{tab:ablation}, our full method substantially improves structural and background preservation compared with this variant while maintaining competitive CLIP similarity. These findings suggest that residual-gated correction helps the editor distinguish regions that require modification from those that should remain unchanged, enabling more targeted edits while preserving the surrounding content.

\medskip
\noindent\textbf{Warm-up.}
As discussed in Sec.~\ref{warm-up}, the warm-up phase allows sufficient residuals to accumulate before enabling correction, helping distinguish regions that require editing from those that should remain unchanged. To evaluate its contribution, we set $K_0 = 0$, enabling correction from the first step. We consider two configurations: \emph{w/o Warm-up} uses the guidance configured for after the warm-up phase throughout all steps, whereas \emph{+ original schedule} retains the original two-stage guidance schedule. As shown in Table~\ref{tab:ablation}, both variants achieve stronger background preservation but lower editing similarity than the full model, indicating that premature correction suppresses intended edits. Their nearly identical results on both backbones suggest that the benefit of the warm-up phase primarily comes from delaying the correction, rather than from the guidance schedule used during this phase.

\medskip
\noindent\textbf{Trajectory-wide versus endpoint-only fidelity.}
As discussed in Sec.~\ref{sec:limitation}, enforcing fidelity only at the endpoint, as in FlowAlign, can be suboptimal. To evaluate the benefit of fidelity throughout the trajectory, we consider an \emph{Endpoint-only fidelity} variant that fixes the terminal time to $S = 1$ in \eqref{eq:soc}, with implementation details provided in the Appendix. As shown in Table~\ref{tab:ablation}, enforcing fidelity throughout the trajectory achieves better structural and background preservation at comparable CLIP similarity. These results suggest that enforcing fidelity throughout the trajectory provides more effective preservation of source content without sacrificing editing capability.

\subsection{Geometric Comparison with FlowAlign Regularization}
\label{sec:ablation_retreat}

The FlowAlign regularizer~\citep{kim2025flowalign} also targets source
consistency. However, it is derived from a quadratic fidelity term
(Sec.~\ref{sec:flowalign}), which makes its correction linear in the
residual. At each step, the \modelname{}~(SD3) pipeline applies two
corrections in sequence: first the FlowAlign regularizer, then the
residual gate (Algorithm~\ref{alg:rage_sd3}). This allows us to
compare the two corrections within the same run.

The question we seek to answer is whether a correction selectively
modifies the trajectory across latent components or merely slows down
the entire editing step. To quantify this distinction, we decompose
each correction $c_k$ at step $k$ into two components: one along the
editing displacement $\Delta_k = (t_{k+1} - t_k)\,(v_p - v_q)$, which
is the update produced by the differential velocity in line~11 of
Algorithm~\ref{alg:rage_sd3}, and the other orthogonal to it:
\begin{equation}\label{eq:retreat_decomposition}
  c_k = -\kappa_k\,\Delta_k + \delta_k, \qquad
  \kappa_k = -\frac{\langle c_k,\,\Delta_k \rangle}
  {\|\Delta_k\|_2^2}.
\end{equation}
Here, $\kappa_k$ is the retreat ratio, indicating what
fraction of the editing step is reversed by the correction, while
$\delta_k$ is the retreat residual, the component that a
uniform retreat cannot produce. We report its magnitude
$\|\delta_k\|_\infty$ and the residual ratio
$\|\delta_k\|_\infty / \|c_k\|_\infty$. For a pure retreat, the
residual ratio is near zero; for a correction that differs
substantially from a retreat, it is of order one.

\begin{figure}[!htb]
  \centering
  \includegraphics[width=0.7\linewidth]{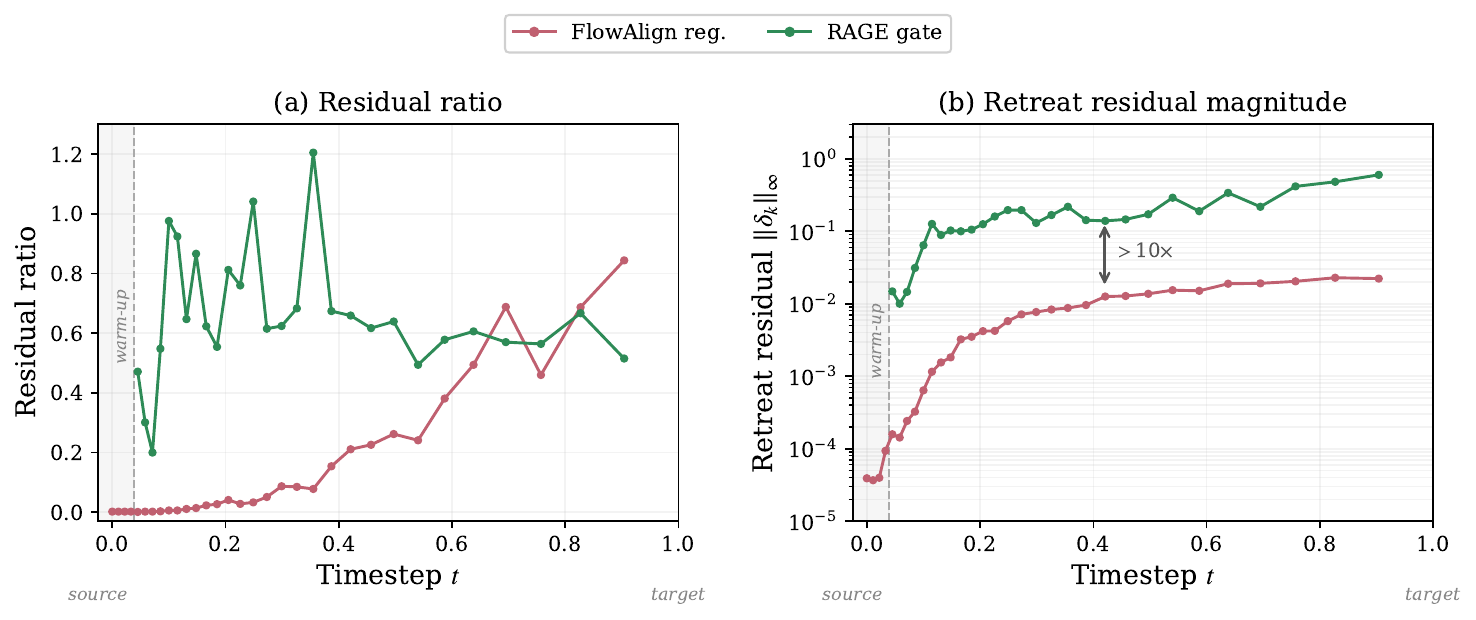}
  \caption{Geometric decomposition of correction terms
    (representative PIE-Bench example, SD3). Gray region: warm-up
    ($K_0\!=\!4$ steps; only FlowAlign active). \textbf{(a)}~Residual ratio: FlowAlign stays near zero in the early phase and rises gradually afterwards; the \modelname{} gate stabilizes at $0.5$--$0.7$ in the late phase (spatially selective). \textbf{(b)}~Retreat residual magnitude (log
    scale): the gated correction exceeds FlowAlign by $>10\times$,
    confirming a structural difference.}
  \label{fig:retreat}
\end{figure}

Fig.~\ref{fig:retreat} shows how these two quantities evolve over time
for a representative SD3 example from PIE-Bench, using the same
configuration as in Table~\ref{tab:quantitative_comparison}. During
the warm-up phase ($K_0 = 4$ steps, shaded in gray), only the
FlowAlign regularizer is active.

\medskip
\noindent\textbf{FlowAlign regularizer.}
In the early phase ($t < 0.2$), the residual ratio of the FlowAlign
correction remains below $0.03$, indicating that the correction is
almost entirely a retreat: it reverses a uniform fraction of the
editing step. The residual ratio increases during the middle phase
but remains below $0.4$ until $t \approx 0.6$, and its retreat
residual remains about an order of magnitude smaller than that of the
gated correction. Thus, for most of the trajectory, the quadratic
fidelity term acts mainly as a trajectory retreat.

\medskip
\noindent\textbf{Gated correction of \modelname{}.}
Once activated, the gated correction maintains a residual ratio of
about $0.5$ or higher for most of the trajectory, stabilizing between
$0.5$ and $0.7$ in the late phase. As measured by the residual ratio, the correction therefore exhibits a substantial deviation from the retreat direction. Its retreat residual is more than ten times that of the
FlowAlign regularizer at every step. This component arises because
the gate applies different correction strengths to different latent
components according to their residuals, which a uniform retreat
cannot achieve.

\subsection{Human Preference Study}

To complement the automatic metrics, we conduct a human preference
study following the pairwise comparison (2AFC) protocol. We randomly
sample 100 images from PIE-Bench as the source pool. Each
comparison draws one image from this pool, pairs it with a randomly
selected baseline on a randomly assigned backbone (FLUX or SD), and presents the source image, the editing instruction, and two
anonymized outputs (ours and one baseline) in randomized left-right
order. Evaluators are asked to select the output that best balances
editing faithfulness and source preservation, or to indicate ``no preference'' if both results are comparable. Each evaluator completes
40 comparisons drawn independently from the shared pool. We
collect responses from 15 evaluators, yielding 600 pairwise judgments
(435 decisive, 160 ties, 5 skipped).

\begin{figure}[!htb]
  \centering
  \includegraphics[width=0.65\linewidth]{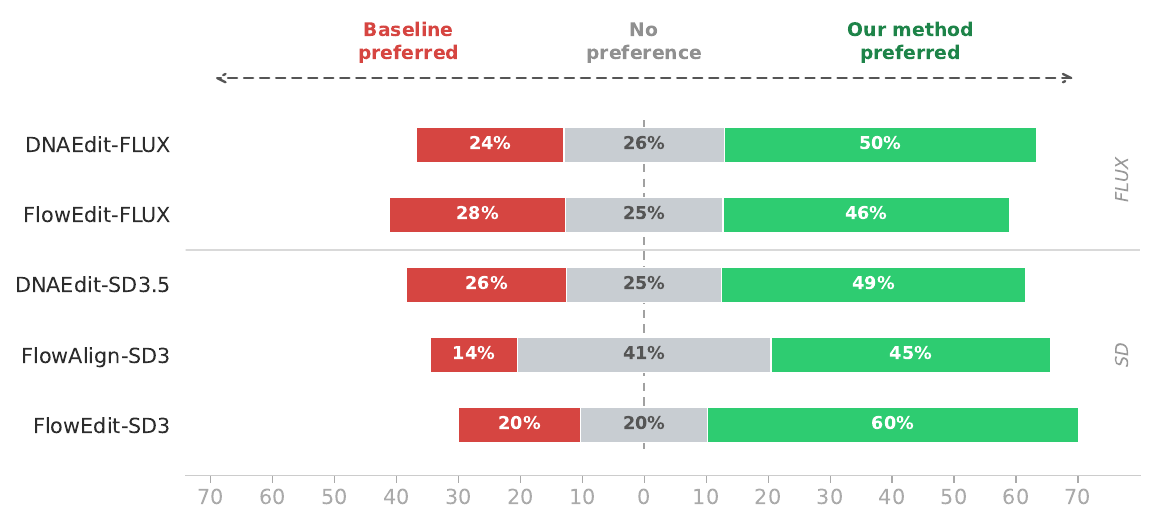}
  \caption{Human preference study on PIE-Bench (pairwise 2AFC,
    15 evaluators, 600 comparisons). Each bar shows the fraction of comparisons in which the baseline is preferred (red), no preference is expressed (gray), or \modelname{} is preferred (green). \modelname{} is
    consistently preferred across all baselines on both backbones.}
  \label{fig:user_study}
\end{figure}

Fig.~\ref{fig:user_study} reports the results. On FLUX,
\modelname{} is preferred over DNAEdit in 50\% of comparisons
(vs.\ 24\% for DNAEdit) and over FlowEdit in 46\% (vs.\ 28\%). On
SD, \modelname{} is preferred over FlowEdit in 60\% of comparisons
(vs.\ 20\%), over DNAEdit in 49\% (vs.\ 26\%), and over FlowAlign
in 45\% (vs.\ 14\%). Across all five baseline pairings on both
backbones, \modelname{} is consistently the most preferred method,
corroborating the quantitative results of
Table~\ref{tab:quantitative_comparison}.

\FloatBarrier

\section{Limitations}

Since the residual gate operates in VAE latent space, \modelname{}
inherits the reconstruction characteristics, and any
biases, of the pretrained backbone. Extending the residual-gating
mechanism to video editing, pixel-space models, and other generative
paradigms beyond rectified flow is a promising direction for future
work.

\section{Conclusion}

We presented \modelname{}, which makes the velocity field of
flow-based image editing source aware. We formulated fidelity control
as a stochastic optimal control problem, which leads to the Residual
Gating Principle: each latent component is pulled back toward the
source with a gain that decays exponentially as its residual
grows. The resulting algorithm needs only one gated interpolation per
sampling step, with no masks, attention maps, retraining, or
architectural changes, and is applied on top of FlowEdit-based editors.
On PIE-Bench, \modelname{} achieves the best overall rank among 17
methods and delivers large, consistent gains in background
preservation over its base editor on both SD3 and FLUX backbones
while maintaining competitive editing quality. We believe the broader
idea of deriving source-aware generative dynamics from optimal control
with selective residual fidelity extends beyond image editing, and we
hope it motivates source-aware formulations in video editing, 3D
editing, and other conditional generation tasks.

\bibliographystyle{iclr2025_conference}
\bibliography{rage_arxiv}	

\end{document}